In-field early disease recognition of potato late blight based on deep learning and proximal hyperspectral imaging

Chao Qi [a, b], Murilo Sandroni [c], Jesper Cairo Westergaard [d], Ea Høegh Riis Sundmark [e], Merethe Bagge [e], Erik Alexandersson [c], Junfeng Gao [a, f, *]

[a] Lincoln Agri-Robotics, Lincoln Institute for Agri-Food Technology, University of Lincoln, Lincoln, UK

[b] College of Engineering, Nanjing Agricultural University, Nanjing 210031, China.

[c] Department of Plant Protection Biology, Swedish University of Agricultural Sciences, Alnarp, Sweden

[d] Department of Plant and Environmental Sciences, University of Copenhagen, Taastrup, Denmark

[e] Danespo Breeding Company, Give, Denmark

[f] Lincoln Centre for Autonomous System, University of Lincoln, Lincoln, UK

**Abstract:** Effective early detection of potato late blight (PLB) is an essential aspect of potato cultivation. However, it is a challenge to detect late blight at an early stage in fields with conventional imaging approaches because of the lack of visual cues displayed at the canopy level. Hyperspectral imaging can, capture spectral signals from a wide range of wavelengths also outside the visual wavelengths. In this context, we propose a deep learning classification architecture for hyperspectral images by combining 2D convolutional neural network (2D-CNN) and 3D-CNN with deep cooperative attention networks (PLB-2D-3D-A). First, 2D-CNN and 3D-CNN are used to extract rich spectral space features, and then the attention mechanism AttentionBlock and SE-ResNet are used to emphasize the salient features in the feature maps and increase the generalization ability of the model. The dataset is built with 15,360 images (64x64x204), cropped from 240 raw images captured in an experimental field with over 20 potato genotypes. The accuracy in the test dataset of 2000 images reached 0.739 in the full band and 0.790 in the specific bands (492nm, 519nm, 560nm, 592nm, 717nm and 765nm). This study shows an encouraging result for early detection of PLB with deep learning and proximal hyperspectral imaging.

## 1. Introduction

Disease monitoring of potatoes is an essential step to improve potato production. Late blight caused by the pathogenic oomycete *Phytophthora infestans* is a devastating disease in potato farming, affecting farmers' income and having a negative impact on the environment relying sometimes on weekly spraying of fungicides (Chen et al., 2021; Zheng et al., 2021). These are expensive and time-consuming to apply (Wang et al., 2021). Hence, accurate early detection of PLB is critical to reduce fungicide applications and, effectively control PLB.

The development of low-cost sensor technologies for computer vision and remote sensing is paving the way for image-based agricultural management and show great potential for automatic detection of crop disease and diagnosis based on different types of images (e.g., RGB images (Su et al., 2021), thermal images (Yang et al., 2021), remote sensing images (Hao et al., 2021)). Our previous work (Gao et al., 2021) has shown great potential to use visual images with deep learning for infield PLB infestation severity evaluation based on the number of recognized lesions. However, it only focuses on advanced stages of PLB development and is unable to recognize pre-symptomatic potato plants infested with *P. infestans*. Hyperspectral imaging (HSI) benefits from capturing many and narrower spectral bands in a continuous spectral range, providing two-dimensional spatial information and rich spectral information in the third dimension.

With the rapid development of hyperspectral remote sensing, many excellent machine learning-based algorithms have been proposed to solve hyperspectral image classification problems (Bhardwaj and Patra, 2018; Hong et al., 2019), such as Support Vector Machine (SVM), polynomial logistic regression, sparse representation, and cooperative representation. Rodriguez et al. (Rodriguez et al., 2021) proposed an Unmanned aerial vehicle (UAV)-based multispectral image detection method for PLB, evaluating the performance of five machine learning algorithms: random forest, gradient augmented classifier, support vector classifier, linear support vector classifier, and k-nearest neighbor classifier to detect PLB with an accuracy of 0.982, 0.895, 0.979, 0.981, respectively. Gold et al. (Gold et al., 2020) used random forest discrimination (RF), partial least squares discriminant analysis (PLS-DA), and normalized difference spectral index to detect regions of late blight occurrence with an accuracy of 70.94%, 71.13%, and 72.23%, respectively. Traditional machine learning methods have some limitations, and the relatively simple mapping structure leads these methods to extract only shallow semantic image features.

To further explore the potential of deep learning in hyperspectral image classification, some algorithms with more advantageous generalization performance have been proposed. Representative models based on deep learning such as convolutional neural networks (CNNs), deep confidence networks, recurrent neural networks, and graph convolutional networks are proposed (Pan et al., 2018; Xu et al., 2018). CNN-based models have been proposed to improve the classification of hyperspectral images of potato leaves because of their excellent performance in many image vision domains (Paoletti et al., 2018). Duarte-Carvajalino et al. (Duarte-Carvajalino et al., 2018) proposed a custom 2D-CNN-based model designed to predict the severity of late blight impact in potato crops using multispectral remote sensing images and state-of-the-art machine learning algorithms, which achieved an accuracy of 0.74. Shi et al. (Shi et al., 2021) proposed a new 3D-CNN deep learning model (CropdocNet) for accurate and automated late blight diagnosis with an accuracy of 94.2%.

Although deep learning-based models can significantly improve the classification performance of hyperspectral images, most existing classification models have the inherent restrictions. They fail to extract sufficient spectral-spatial correlation information. The 2D-CNN-based deep network structure only utilizes spatial information, missing the important information from spectral signals. Similarly, the 3D-CNN-based network model performs poorly in many spectral bands to classify similar texture classes. The purpose of the method presented here is to accurately detect PLB at an early stage. During the early stage of late blight, during the biotrophic life stage of *P. infestans,* the degree of leaf infectioncannot be judged by human inspection, thus the capability of extracting spectral-spatial features is critical. To tackle this, we propose a deep learning model combining 2D-CNN and 3D-CNN to make full use of the respective advantages of the two through customized feature extraction for the early detection PLB. The main contributions of this paper are as follows:

1. A customized deep learning structure designed for early detection of PLB by combining 2D-CNN and 3D-CNN as well as attention networks.

2. Extracting the important bands for PLB classification and validating its classification effectiveness compared to classification with full wavelengths. 3. Demonstrating the superiority of model PLB-2D-3D-A is verified by comparing with traditional machine learning methods and deep learning methods (2D-CNN and 3D-CNN).

The rest of the paper is organized as follows: Section 2 describes the dataset and the proposed model, Section 3 presents the experimental results, Section 4 provides a discussion and highlights future work, and Section 5 summarizes this work.

## 2. Materials and methods

### *2.1. Dataset*

The data were collected on July 9, July 13, July 15, and July 18, 2020 as part of the Danespo field trial outside of Give, Denmark (N 55.800586, E 9.223748) and contained 20 different potato genotypes varying in susceptibility to PLB (four cultivars and sixteen breeding lines). Hyperspectral images with a resolution of 512*512*204 and a band of 204 were collected using a handheld hyperspectral camera, Specim IQ, from Specim (Oulu, Finland). Potatoplants from the infection rowswere inoculated with *P. infestans* on July 7 2020, from where the pathogen was expected to spread over the studied plots. Images collected on the third, seventh, ninth and twelfth day after inoculation were classified as one category, respectively. Among them, the images collected in the third, seventh and ninth days could not be discerned by the naked eye as no disease spots were displayed, and we defined them as very early stage of late blight. While few of the images in the twelfth day could be discerned by the naked eye as disease spots, we still consider them as early PLB. To reduce the computational burden, each collected raw image (512*512*204) were cropped into 64 subimages (64*64*204) to scale up training samples and reduce computation burden. To ensure the integrity of the spatial information, we did not pre-process the channel information. The number of images per genotype is 12, thus the total number of data set is 15360 (64*12*20) and the number of each category is 3840. The examples of original images in each category are shown in Fig. 1.

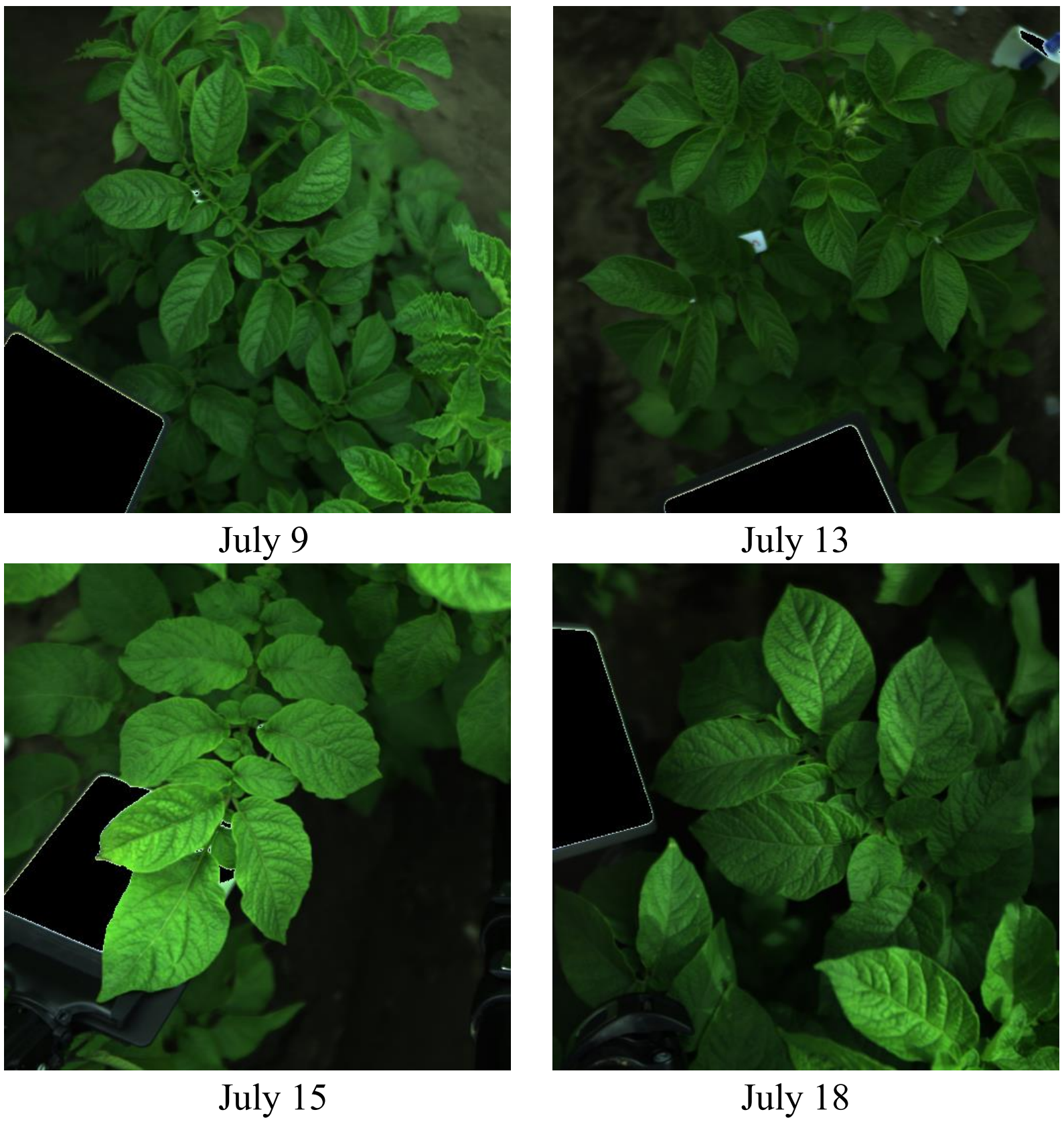

**Fig. 1.** Examples of the infected potatoes at four different dates, displaying with RGB bands.

*2.2. Radiation calibrations*

Radiation calibration mainly includes dark current removal from the original data radiation intensity, gain calibration of each spectral data channel and wavelength revision, etc. The Equation is as follows

$$TR = \frac{DNo}{DNr} \times RP_r \quad (1)$$

Where DNo represents the pixel value of the object. DNr stands for the pixel value of the reference plate. RPr stands for the reflectance of the reference plate, obtained by prior calibration in the laboratory.

The calculated object reflectance TR is used as one of the basic parameters for calculating the radiance value of this image element at the sensor pupil entry using the radiative transfer equation. The multiple spectral reflectance curves for each sample point from the processing are averaged and the average value is taken as the spectrum for that sample point. The instability of the measurement is removed by multiple measurements on the same object and the results are more representative and credible.

Assuming that the calibration is linear, we have

$$L = a^{*}DC + b \quad (2)$$

$$D = a^{*}DC_0 + b \quad (3)$$

where L refers to the radiance of the image at the sensor pupil $(W/(m^2 \cdot sr \cdot um))$ calculated using the radiative transfer model. D indicates the irradiance of the corresponding image in the dark current case $(W/(m^2 \cdot sr \cdot um))$. DC represents the grey scale value of the image element on the corresponding remote sensing image. $DC_0$ describes the dark current value of the image element on the corresponding remote sensing image. a and b are the calibration coefficients to be found. In the dark current case, D should be zero, and the calibration coefficients a and b can be found by combining Equations (2) and (3). The images calibrated by radiation are shown in Fig. 2.

$$a = L/(DC - DC_0) \quad (4)$$

$$b = -a \cdot DC_0 \quad (5)$$

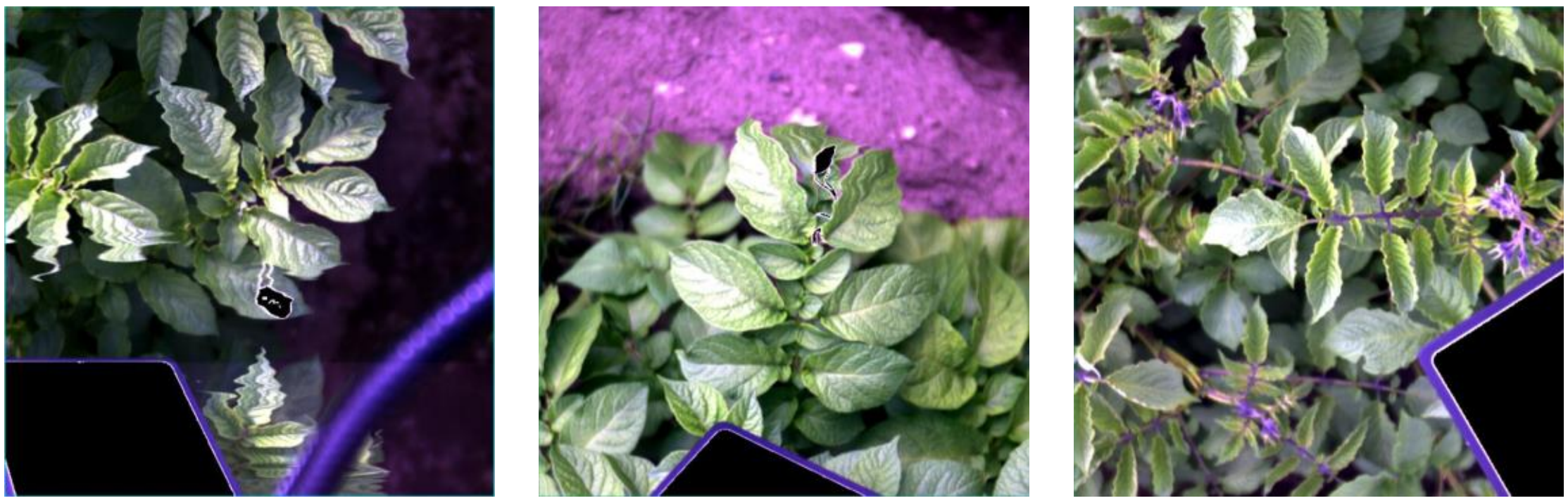

**Fig. 2.** Some images after radiometric calibration.

*2.3. Background removal*

The purpose of background removal is to remove information from hyperspectral images of objects besides potato leaves that interfere with detecting and evaluating late blight in the experimental area. The noisy objects in the images includes calibration plates, marker plates, tripods, etc. The images could be collected in the varying illumination, so it is difficult to remove these backgrounds by selecting a specific threshold. Through the experiment, we learnt that when the standard deviation of a pixel in an image is greater than one half of the pixel mean, the pixel is set to NaN (in mathematical terms, it means an unrepresentable number that does not participate in calculations). When the standard deviation of a pixel in an image is less than one-half of the pixel mean, the original pixel value is kept unchanged with the following formula.

$$\begin{cases} \sqrt{\frac{1}{N}\sum_{i=1}^{N}(x_i-\mu)^2} \geqslant \frac{N}{2}, 0 \\ \sqrt{\frac{1}{N}\sum_{i=1}^{N}(x_i-\mu)^2} < \frac{N}{2}, x_i \end{cases} \tag{6}$$

Where $N$ represents the total number of pixels, $x_i$ stands for the pixel value and $\mu$ indicates the pixel average.

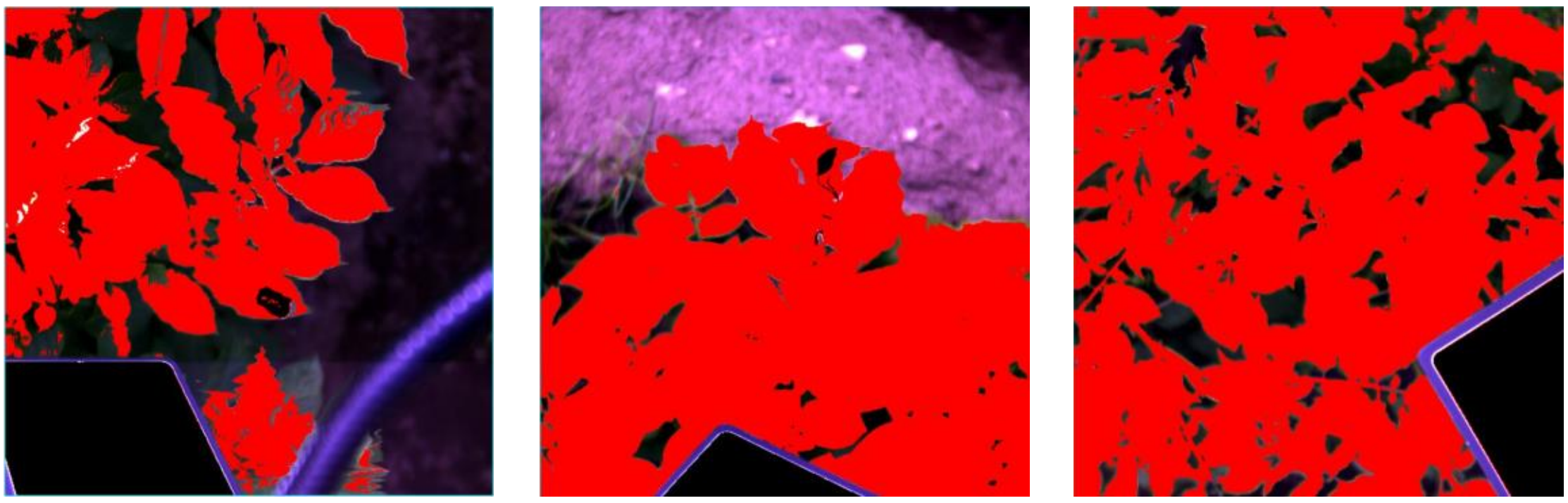

**Fig. 3.** Results of background removal.

### *2.4. Band Screening*

Although hyperspectral images provide a great deal of finer feature information, not all the images from each band could provide useful information with the increase of spectral wavelengths. Redundancy in hyperspectral imagery includes spatial and spectral redundancy. Spatial redundancy arises at the same band, as the greyscales of sampled points on the surface of the same feature are typically spatially coherent with each other, and feature greyscales based on discrete pixel sampling do not take full advantage of this feature. Spectral redundancy arises because the high spectral resolution and high data dimensionality of hyperspectral images allow information in one band of the image to be partially or fully predicted by other bands. Specifically, our dataset captures 204 bands, leading to that the parameter number to be processed reaches about 50 million for each image. The redundant data brings a heavy computational burden, prone to slow or even failed network training. Therefore, it is necessary to verify whether screening out specific bands has an impact on the detection results. We use spectral first-order differentiation and spectral second-order differentiation to conduct band screening separately, and the final results are intersected to improve the quality of the band screening. The principles of spectral first order differentiation and spectral second order differentiation are as follows:

The derivative of a function is defined as:

$$f'(x) = \lim_{h\to 0}\frac{f(x+h)-f(x)}{h} \tag{7}$$

While the $h$ is small enough, we can use a centered difference formula to approximate the derivative:

$$f'(x_i) \approx \frac{f(x_i+h)-f(x_i-h)}{2h} \quad (8)$$

In practice, Origin treats discrete data by the transform of the centered difference formula, and calculates the derivative at point $P_i$ by taking the average of the slopes between the point and its two closest neighbors.

The derivative function applied to discrete data points can therefore be written:

$$f'(x_i) = \frac{1}{2}\left(\frac{y_{i+1}-y_i}{x_{i+1}-x_i} + \frac{y_i-y_{i-1}}{x_i-x_{i-1}}\right) \quad (9)$$

When smooth option is chosen in differentiate, and X data is evenly spaced, Savitzky-Golay method will be used to calculate the derivatives.

First perform a polynomial regression on the data points in the moving window. The polynomial value at position x can be calculated as:

$$f(x) = a_n x^n + a_{n-1}x^{n-1} + a_{n-2}x^{n-2} + \cdots + a_1 x + a_0 \quad (10)$$

where n is the polynomial order, and $a_i, i = 0 \ldots n$ are fitted coefficients.

And first order derivative at position x is:

$$f'(x) = na_n x^{n-1} + (n-1)a_{n-1}x^{n-2} + \cdots + a_1 \quad (11)$$

Second order derivative at position x is:

$$f''(x) = n(n-1)a_n x^{n-2} + (n-1)(n-2)a_{n-1}x^{n-3} + \cdots + a_2 \quad (12)$$

### *2.5. Classification model*

#### *2.5.1. PLB-2D-3D-A*

First, the pre-processed image (512*512*204) is uniformly sliced into 64*64*204 hyperspectral images using 3D convolution. To avoid the loss of spatial-spectral information, we only slice for pixel information instead of channel information. Second, to better investigate spatial-spectral features, this paper designed a sub-structure containing 2D-CNN and 3D-CNN. In this sub-structure, spatial feature information is extracted by 2D-CNN and spectral-spatial context information is obtained by 3D-CNN, and BN and RELU are embedded after each convolution operation. Moreover, the extracted features are combined with two attention mechanisms, AttentionBlock and SE-ResNet, to fully extract spectral-spatial features and emphasize saliency features. Finally, the identifiable spatial-spectral features are fed into a $1 \times 1$ convolutional layer to facilitate classification. The PLB-2D-3D-A structure is shown in Fig. 4. The experiments were conducted on a server with NVIDIA Tesla V100, CUDA 11.2, and Ubuntu 16. During the training process, the key hyperparameters were set as follows: dropout=0.4, epoch=60, batch size=16, learning rate=$2e^{-4}$, and the optimizer was Adam.

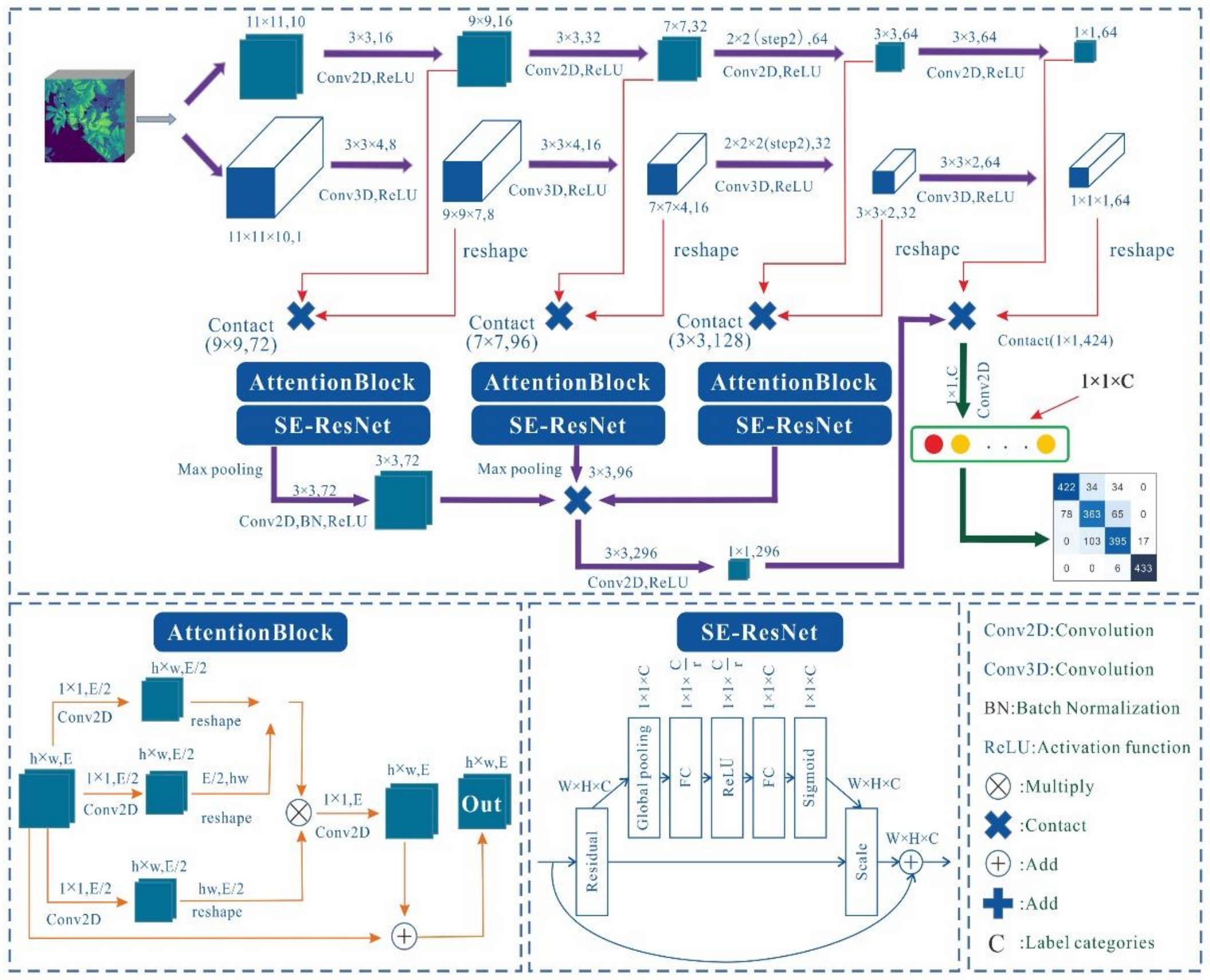

**Fig. 4.** The PLB-2D-3D-A structure.

*2.5.2. Spatial feature extraction*

To extract feature maps with sufficient spatial-spectral contextual information, two-dimensional and three-dimensional convolution operations were designed in the PLB-2D-3D-A model. The following describes how to combine the 2D CNN and 3D CNN so that the feature maps comprise rich spatial-spectral correlation information. Before passing into the deep network, a neighborhood block $P \in \mathbb{R}^{S \times S \times B}$ is created by selecting a neighborhood square of size $S \times S$ around the center pixel from $I$, with the spatial location of the center at $(\alpha, \beta)$ and the number of bands $B$. In the experiments, $S$ is set to 11 and $B$ to 10.

For spatial correlation feature extraction, 2D-CNN operations are used as the basic unit for spatial feature extraction. The spatial extraction region contains four 2D-CNN $(\mathrm{cov2}D_\{i\}_{i=1}^{4})$ layers to capture feature maps of various spatial dimensions. After the conv2D_1、conv2D_2 and conv2D_4 operations, 3×3 convolution kernels with (1,1) downsampling steps are applied. conv2D_3 uses 2×2 convolution kernels with (2,2) steps.

For spectral relevant feature extraction, 3D convolution operations are employed to capture the correlation of the spectral dimensions. As shown in Figure 1, four 3D convolutional layers $(\mathrm{conv3D}_\{i\}_{i=1}^{4})$ operations are deployed to acquire spectral features of various depths. In conv3D_1, conv3D_2 and conv3D_4, the three 3D convolution kernels are 3×3×4, 3×3×4 and 3×3×2, respectively, where each convolution kernel has a downsampling step of (1,1,1). conv3D_3 contains convolution kernels of 2×2×2 and a step of (2,2,2).

The first three spatially relevant feature maps are fused with the spatial-spectral correlated feature maps so that the fused feature maps comprise rich relevant information. The fourth spatially correlated feature map is fused with the spectral-spatially correlated feature map to assist classification.

### 2.5.3. AttenionBlock

Based on the later experimental results, it can be observed that the PLB-2D-3D-A model including the attention module AttenionBlock gives superior classification results. Further investigation identified the following steps explains the effectiveness of AttenionBlock from both theoretical and experimental perspectives. In the proposed model, AttenionBlock is utilized to emphasize the relevant information in the fused spectral space feature map. First, all the pixels in the feature map can be represented as $X_{in} = \{x_1, x_2, \dots, x_N\}$, where $N$ denotes the number of pixels. Each pixel $x_i$ is represented as an $E$ dimensional vector, where $E$ refers to the channel number of the feature map. The output of this module is $O_{\text{out}} = \{o_1, o_2, \dots, o_N\}$. The similarity between each two pixels $x_i$ and $x_j$ can be expressed as:

$$o_i = \text{softmax}\left(\phi(x_i)^T \varphi(x_j)\right) g(x_j)$$

$$= \frac{1}{\Sigma_{\forall j} exp^{\exp\left(\phi(x_i)^T \varphi(x_j)\right)}} \left(\phi(x_i)^T \varphi(x_j)\right) g(x_j) \quad (13)$$

Where $\phi(\cdot)$, $\varphi(\cdot)$and g$(\cdot)$ all represent a Conv2D in Figure 4-2 with a convolution kernel size of 1×1 and several convolution layers of $E/2$. Then, through two matrix multiplication operations and one regular softmax operation, the difficulty of the feature map obtained is $h \times w \times E/2$ (where $h$ and $w$ represent the width and height of $X_{in}$, respectively). The latitude of the feature map obtained using $E$ Conv2D with a spatial kernel size of 1×1 to do the spatial convolution operation on the feature map is $h \times w \times E$. Finally, the obtained feature map is added with $X_{in}$ to get the output $O_{out}$.

In contrast to convolution and pooling operations, AttenionBlock considers the relationship between distant pixels with the weights of all positions in the feature map. AttenionBlock focuses on the relevance between pixels in the whole feature map, whereas the convolution and pooling operations only concentrate on pixels in the spatial range of the convolution kernel size. The spatial size of the fused feature map is 9 × 9, 7 × 7, and 3 × 3. Since the spatial range is not very large, thus using AttenionBlock will only incur little computational complexity. Based on the above analysis, the effectiveness of AttenionBlock in improving classification performance is theoretically verified.

### *2.5.4. SE-ResNet*

For CNN networks, the core computation is the convolution operator, which learns from the input data to the feature maps by employing a series of convolution kernels. In essence, convolution is the fusion of features over a local region, which includes spatially ($H$ and $W$ dimensions) as well as inter-channel (C dimension) fusion of features. For convolution operations, a large part of the work is to improve the perceptual field, i.e., to spatially fuse more feature fusions, or to extract multi-scale spatial information, such as the multi-branch structure of Inception networks. For feature fusion in channel dimension, the convolution operation fuses all channels of the input feature map by default. The innovation of the SENet network (Li et al., 2021) is to focus on the relationship between channels, hoping that the model can automatically learn the importance of different channel features. For this purpose, SENet proposes the Squeeze-and-Excitation (SE) module, as shown in the following figure.

The SE module first performs the Squeeze operation on the feature map resulting from the convolution to acquire the global features at the channel level, followed by an excitation operation on the global features to learn the relationship between each channel, while obtaining the weights of the different channels, and finally multiplies the original feature map to obtain the final features. Essentially, the SE module is doing the attention or the rating operation on the channel dimension. This attention mechanism allows the model to pay more attention to the most informative channel features and suppress those unimportant channel features. Another point is that the SE module is generic, which means it can be integrated into existing network architectures.

The SE module mainly comprises two operations, Squeeze and Excitation, which can be adapted to any mapping. Take convolution as an example, the convolution kernel is $V = [v_1, v_2, \ldots, v_C]$, where $v_C$ denotes the $C_{\text{th}}$ convolution kernel. The output $U = [u_1, u_2, \ldots, u_C]$.

$$u_c = v_c * X = \sum_{s=1}^{C'} v_c^s * x^s \tag{14}$$

Where * stands for the convolution operation, and $v_c^s$ stands for a 2D convolution kernel of an *s* channel, whose input is the spatial features on a channel, and it learns the feature spatial relations, but since the sum is done on the convolution results of each channel, the channel feature relations are mixed with the spatial relations learned by the convolution kernel. The SE module is designed to abstract away this mixing, so that the model can directly learn the channel feature relations.

Since convolution only operates in a local space, it is difficult for *U* to obtain enough information to extract the relationship between channels, which is more serious for the front layers of the network, because the perceptual field is relatively small. SENet proposes the Squeeze operation, which encodes the entire spatial feature on a channel as a global feature, using global average pooling to achieve.

$$z_c = F_{sq}(u_c) = \frac{1}{H \times W} \sum_{i=1}^{H} \sum_{j=1}^{W} u_c(i,j), z \in R^C \tag{15}$$

After the Squeeze operation retrieves the global description of the features, we need another operation to capture the relationships between channels. This operation needs to satisfy two criteria: firstly, it has to be flexible, and able to learn the nonlinear relationships between the individual channels; secondly, the learned relationships are not mutually exclusive, because here multi-channel features are allowed instead of the one-hot form. Based on this, the gating mechanism in sigmoid form is used here:

$$s = F_{ex}(z, W) = \sigma(g(z, W)) = \sigma\big(W_2 \mathrm{Re}LU(W_1 z)\big) \tag{16}$$

Where $W_1 \in R^{\frac{C}{r} \times C}$ and $W_2 \in R^{C \times \frac{C}{r}}$. To reduce the model complexity as well as to improve the generalization ability, a bottleneck structure containing two fully connected layers is adopted here, where the first FC layer plays the role of dimensionality reduction, and the dimensionality reduction factor of *r* is a hyperparameter, and then ReLU activation is applied. The final FC layer restores the original dimensionality. Finally, the learned activation values (sigmoid activation, values 0 to 1) of each channel are multiplied by the original features on *U*:

$$\tilde{x}c = Fscale(u_c, s_c) = s_c \cdot u_c \tag{17}$$

In fact, the whole operation can be seen as learning the weight coefficients of each channel, thus making the model more discriminative of the features of each channel, which should also be considered an attention mechanism. SE module applied in ResNet (Zhang et al., 2021), the model parameters and the amount of computation will increase, here take SE-ResNet-50 as an example, for the increase of model parameters is

$$\frac{2}{r}\sum_{s=1}^{S} N_s \cdot C_s^2 \quad (18)$$

where $r$ denotes the number of descending coefficients, $S$ indicates the number of stages, $C_s$ refers to the number of channels in the sth stage, and $N_s$ stands for the amount of duplicate blocks in the sth stage. When $r$=16, SE-ResNet-50 only increases the number of parameters by about 10%. However, the computational volume (GFLOPS) is increased by less than 1%.

*2.6. RF, 2D-CNN and 3D-CNN*

Random Forest (RF) (Li et al., 2019) is a machine learning algorithm that combines hundreds of decision trees, where each tree depends on the values of independently sampled random vectors (Breiman, 2001). Predictions are aggregated for classification or averaged for regression by majority voting on the predictions of the set. Since RF is a non-parametric method, it does not require values to follow a specific statistical distribution.

PLB-2D-3D-A is a classification algorithm based on 2D-CNN in cooperation with 3D-CNN, we split A into a 2D-CNN based classification algorithm and a 3D-CNN based classification algorithm, in the splitting process, we only change the feature extraction part, that is, we keep the whole network having an only 2D-CNN structure or 3D-CNN structure, the specific structure is shown in Fig. 5.

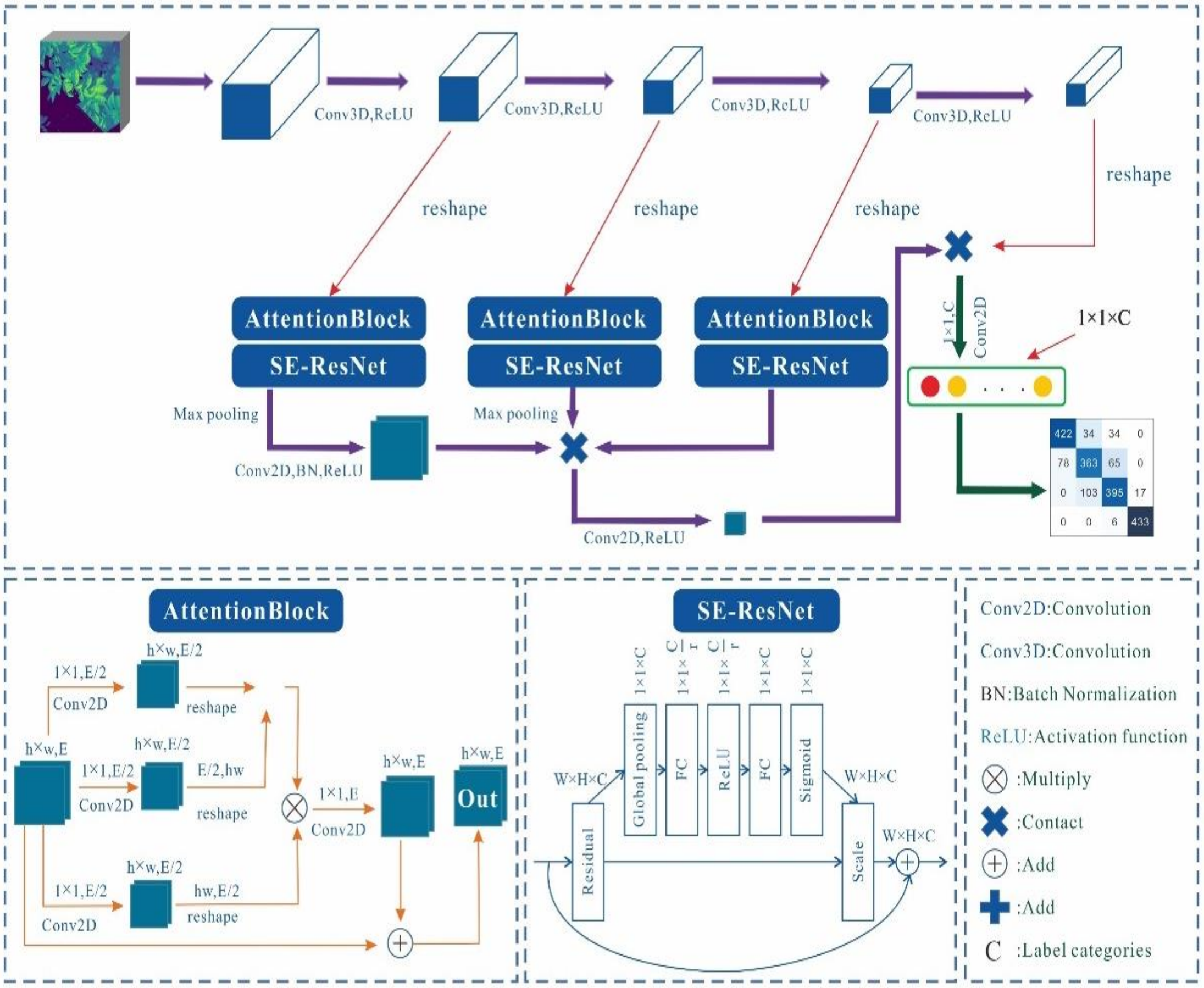

2D-CNN structure

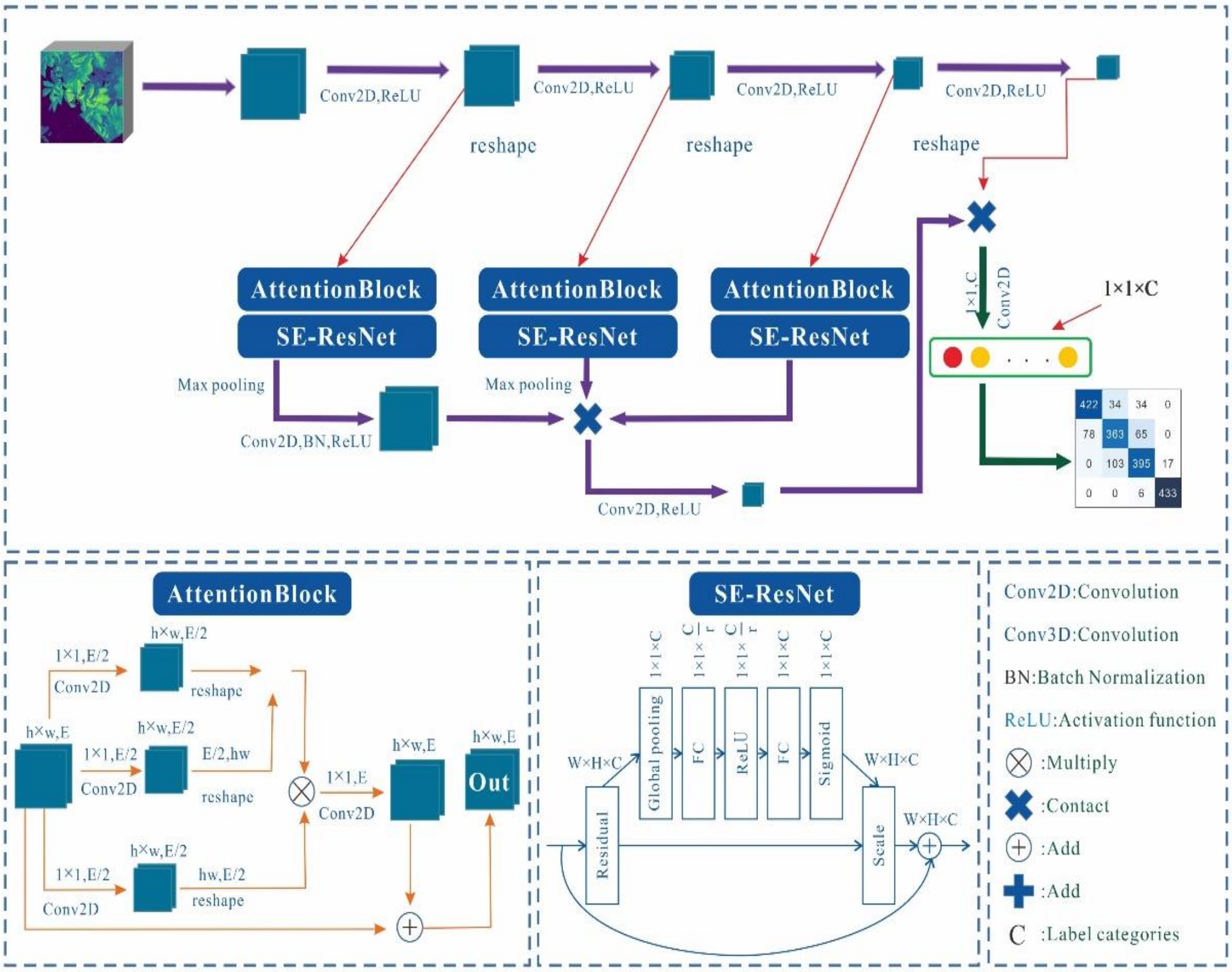

3D-CNN structure

**Fig. 5.** Customized 2D-CNN structure and 3D-CNN structure.

*2.7. Evaluation metrics*

Based on the confusion matrix, we obtained the following metrics: Accuracy, Precision, Recall and F1 Score (Jozdani and Chen, 2020). the calculation formula is as follows:

$$\text{Accuracy} = \frac{TP+TN}{TP+FN+FP+TN} \tag{19}$$

$$\text{Precision} = \frac{TP}{TP+FP} \tag{20}$$

$$\text{Recall} = \frac{TP}{TP+FN} \tag{21}$$

$$\text{Accuracy} = \frac{2\times\text{Precision}\times\text{Recall}}{\text{Precision}+\text{Recall}} \tag{22}$$

Where TP, FP, TN and FN are the true positive, false positive, true negative and false negative samples, respectively.

*2.8. Experimental setup*

We selected full-band samples and six specific bands from the band screening in Section 2.4 as input samples, respectively, to verify the classification effectiveness of model PLB-2D-3D-A for the early PLB. Not only that, to investigate the detection performance of PLB-2D-3D-A in-depth, we compare the training results with the performance of three classification models (RF, 2D-CNN and 3D-CNN).

## 3. Results

We investigated specific bands of 20 different infected potato genotypes and Fig. 6 illustrates the spectral curves of infected potato genotypes. According to the methodology in Section 2.4, some results are shown in Fig. 7, where the peaks and troughs represent specific bands. Taking the intersection of all specific bands, we screened six important bands (492nm, 519nm, 560nm, 592nm, 717nm and 765nm).

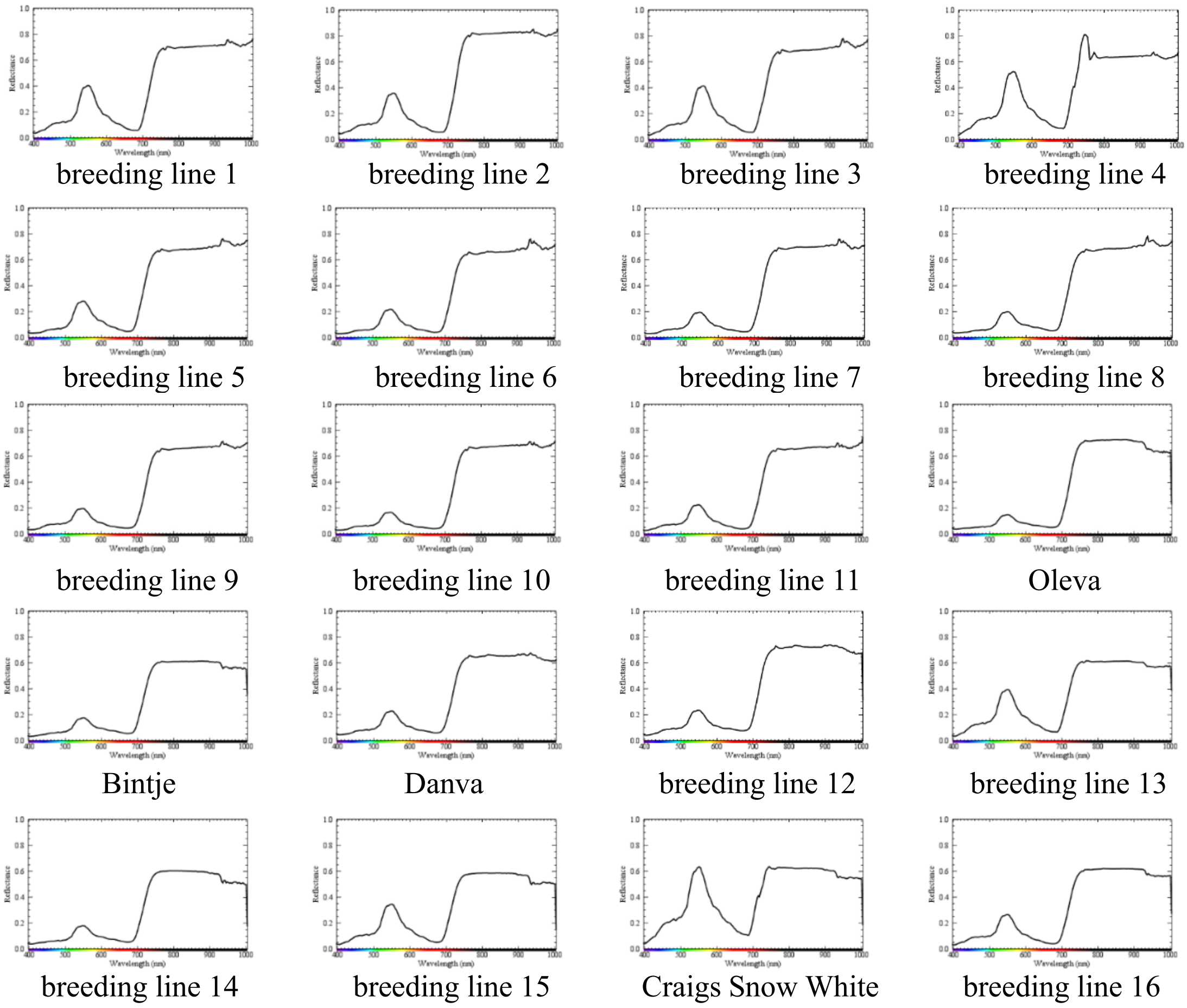

**Fig. 6.** Results of band screening.

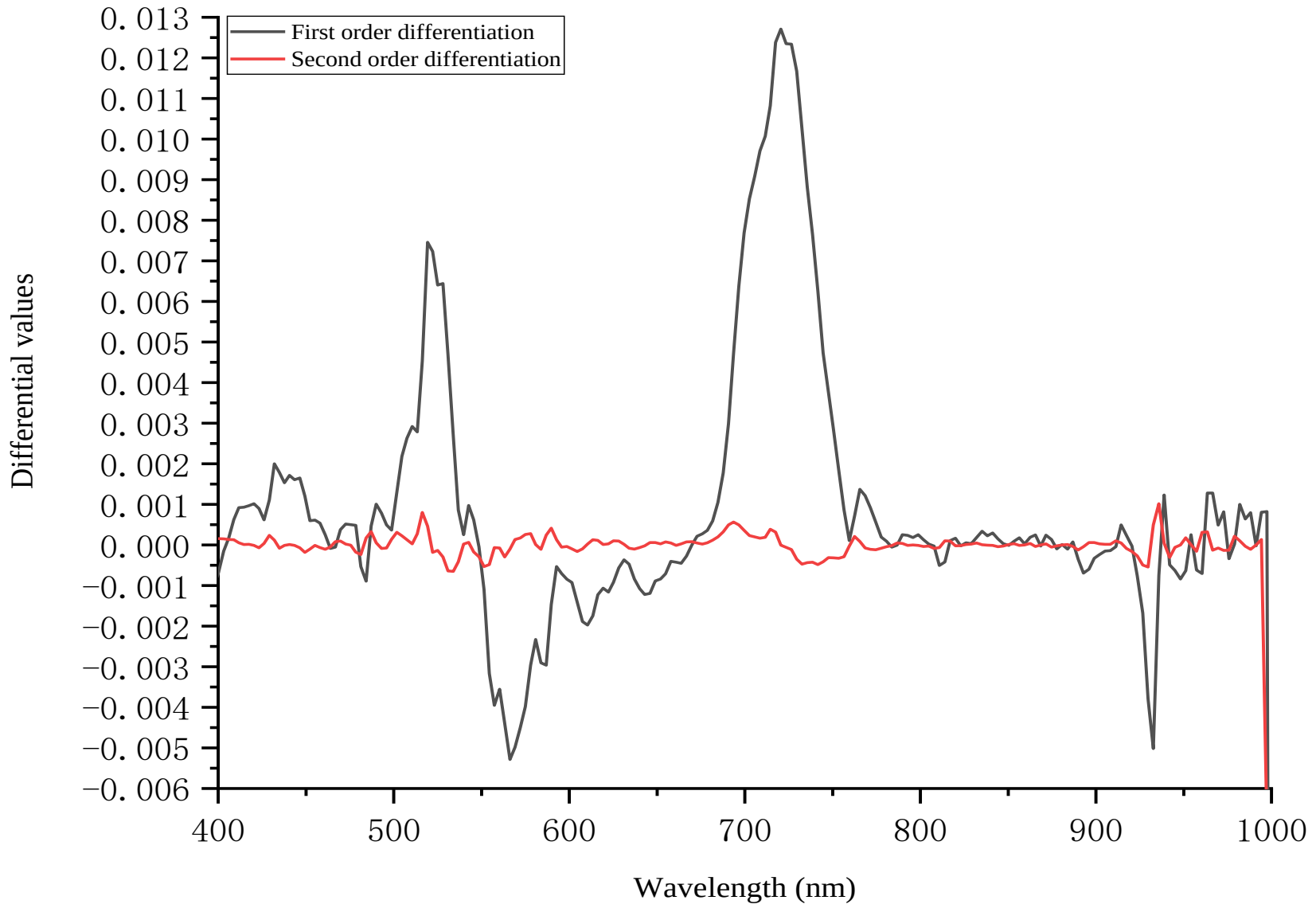

**Fig. 7.** First and second order differentiations for spectral feature selection.

Table 1 shows the classification results of the four models under the full bands, and Fig. 8 shows the corresponding confusion matrix. It can be seen that PLB-2D-3D-A has a significant performance advantage, reaching an accuracy of 0.739. As expected with increased infection, the fourth class (July 18) has the best classification results, with precision, recall and F1 can reach 0.838, 0.892 and 0.864, respectively. Surprisingly, the first class (July 9) was second only to the fourth in terms of classification, with precision, recall and $F_1$ reaching 0.764, 0.830 and 0.796, respectively. The inoculation was on July 7, which means that on the third day of infection, the handheld spectrometer was already able to capture the reflectance changes of the potato leaves. However, the classification performance of the second class (July 13) and the third class (July 15) is somewhat disappointing, especially for the second class, where precision, recall and $F_1$ only reach 0.632, 0.623 and 0.627, respectively, as seen from the confusion matrix in Fig. 8. Not only the PLB-2D-3D-A model, but also the RF, 2D-CNN and 3D-CNN have similar issues, with a large number of features entangled in the second and third classes. The RF achieves the lowest accuracy (0.683), probably because it fails to learn high-level complex image features in-depth. 3D-CNN has a slightly better performance than 2D-CNN, showing that the accurate acquisition of spatial information plays a key role in analyzing hyperspectral images.

**Table 1.** The classification results of the four models under the full bands.

| Models | | July9 | | | July13 | | | July15 | | | July18 | | |
|---|---|---|---|---|---|---|---|---|---|---|---|---|---|
| | Accuracy | precision | recall | F1 | precision | recall | F1 | precision | recall | F1 | precision | recall | F1 |
| RF | 0.683 | 0.696 | 0.825 | 0.755 | 0.576 | 0.535 | 0.555 | 0.680 | 0.571 | 0.621 | 0.778 | 0.874 | 0.823 |
| 2D-CNN | 0.705 | 0.722 | 0.830 | 0.772 | 0.592 | 0.566 | 0.579 | 0.698 | 0.596 | 0.643 | 0.806 | 0.884 | 0.843 |
| 3D-CNN | 0.717 | 0.730 | 0.820 | 0.772 | 0.604 | 0.584 | 0.594 | 0.710 | 0.616 | 0.660 | 0.822 | 0.890 | 0.855 |
| Ours | 0.739 | 0.764 | 0.830 | 0.796 | 0.632 | 0.623 | 0.627 | 0.722 | 0.638 | 0.677 | 0.838 | 0.892 | 0.864 |

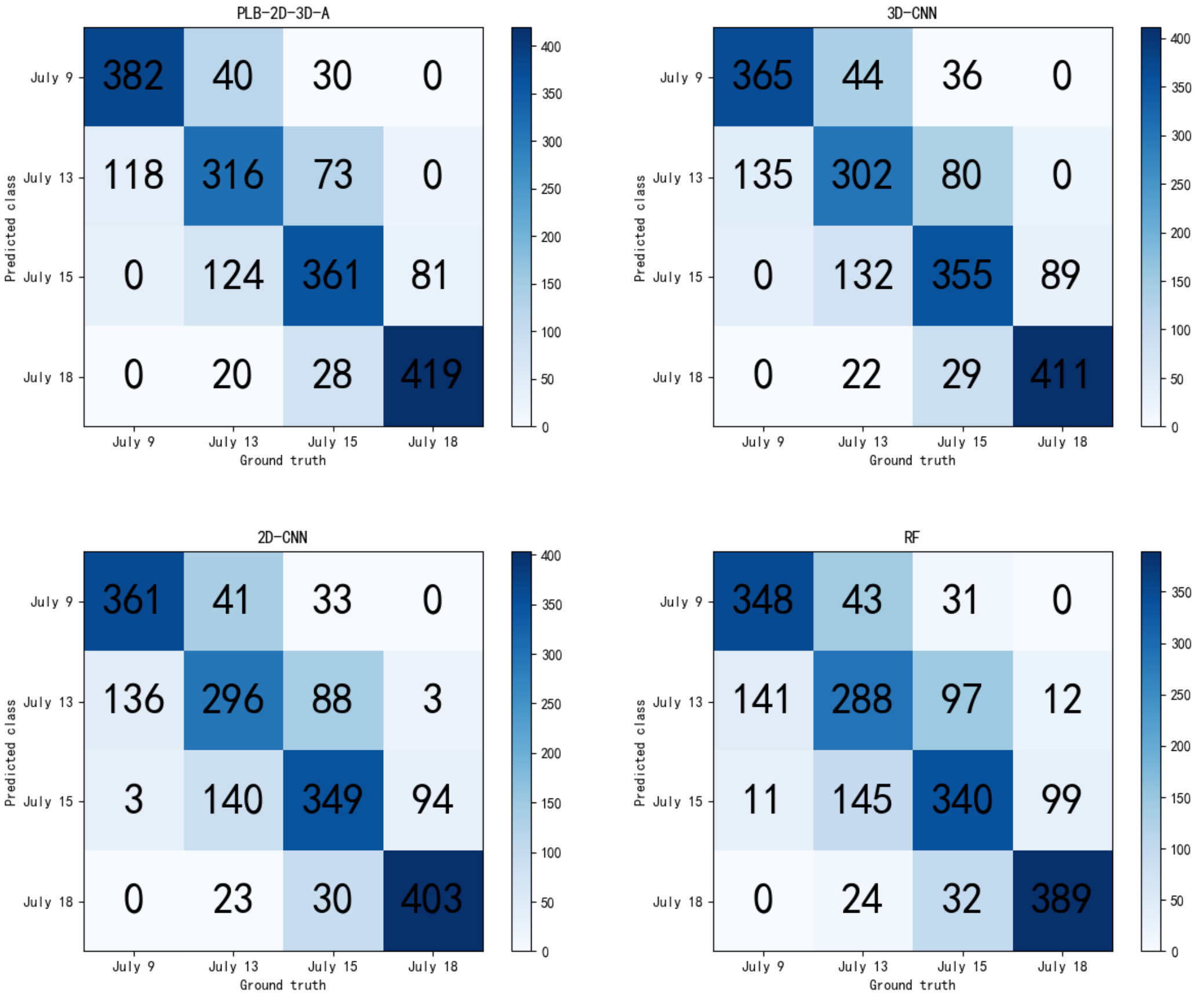

**Fig. 8.** Confusion matrix results of the four models at full bands.

Table 2 shows the classification results of the four models under six specific bands, and Fig. 9 shows the corresponding confusion matrices. When we select specific bands with high reflectance, the overall performance of the model is slightly improved, with RF, 2D-CNN, 3D-CNN and PLB-2D-3D-A Accuracy improved by 5.56%, 6.38%, 6.83% and 6.9%, respectively. It can be seen that the selection of specific bands is effective in improving classification performance.

**Table 2.** The classification results of the four models under the specific bands.

| Models | | July9 | | | July13 | | | July15 | | | July18 | | |
|---|---|---|---|---|---|---|---|---|---|---|---|---|---|
| | Accuracy | precision | recall | F1 | precision | recall | F1 | precision | recall | F1 | precision | recall | F1 |
| RF | 0.721 | 0.734 | 0.889 | 0.804 | 0.674 | 0.589 | 0.629 | 0.704 | 0.584 | 0.638 | 0.772 | 0.937 | 0.847 |
| 2D-CNN | 0.750 | 0.762 | 0.884 | 0.829 | 0.686 | 0.616 | 0.649 | 0.734 | 0.637 | 0.682 | 0.818 | 0.938 | 0.874 |
| 3D-CNN | 0.766 | 0.774 | 0.882 | 0.824 | 0.704 | 0.648 | 0.675 | 0.748 | 0.658 | 0.700 | 0.836 | 0.929 | 0.880 |
| Ours | 0.790 | 0.796 | 0.888 | 0.839 | 0.720 | 0.679 | 0.699 | 0.764 | 0.700 | 0.731 | 0.878 | 0.922 | 0.899 |

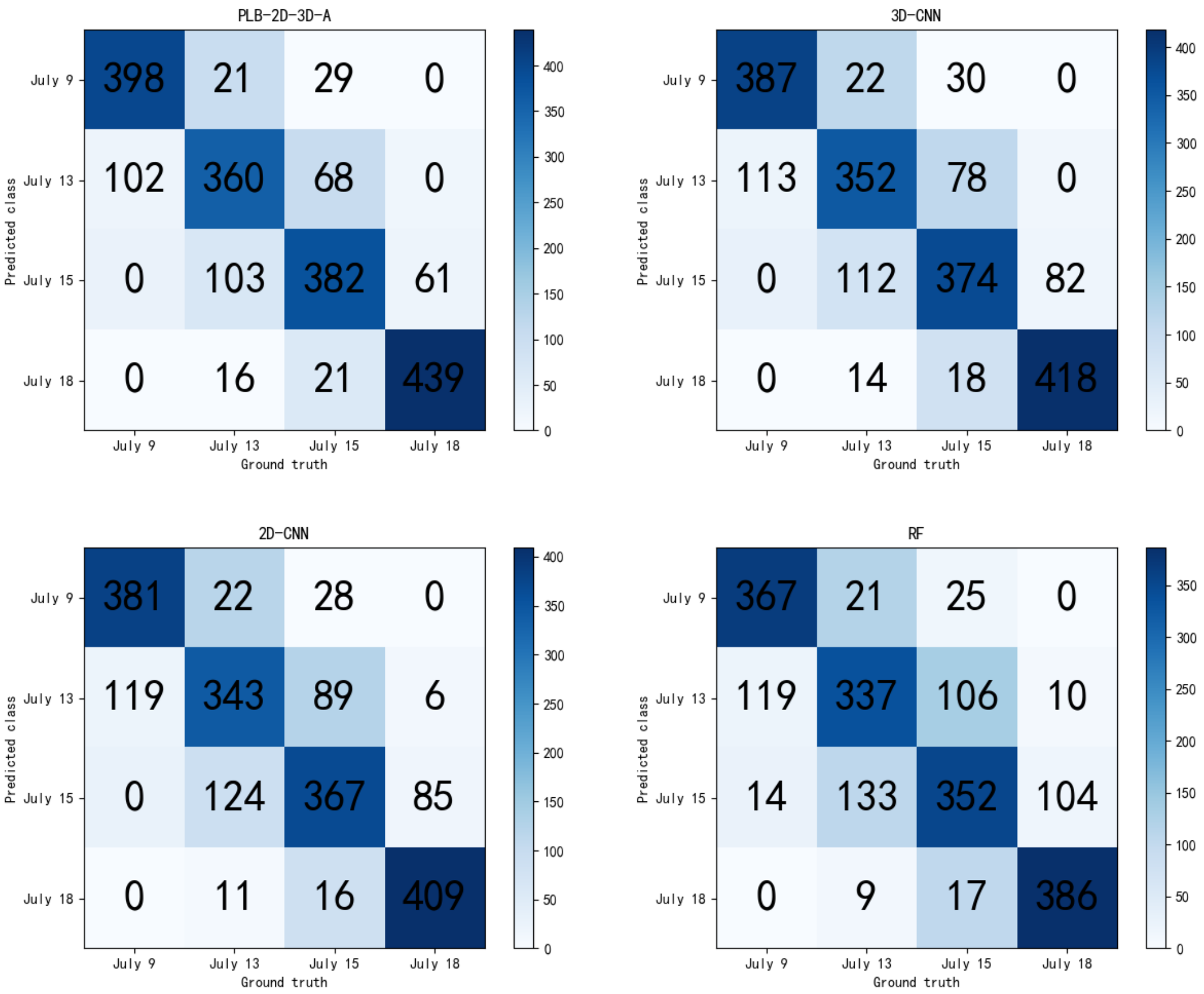

**Fig. 9.** Confusion matrix results of the four models at specific bands.

To investigate the effect of attentional mechanisms on the proposed PLB-2D-3D-A, we designed ablation experiments to validate the AttenionBlock module and the SE-ResNet module. Specifically, we removed the AttenionBlock module and the SE-ResNet module in turn and tested them at full and specific bands, respectively, with the experimental setup remaining consistent with the above experiments. The detailed data of the results are shown in Table 3 and Table 4, and the confusion matrices in the test dataset are shown in Fig. 10 and Fig. 11.

**Table 3.** Ablation experiments at full bands.

| Models | | July9 | | | July13 | | | July15 | | | July18 | | |
|---|---|---|---|---|---|---|---|---|---|---|---|---|---|
| | Accuracy | precision | recall | F1 | precision | recall | F1 | precision | recall | F1 | precision | recall | F1 |
| Ours-A-S | 0.700 | 0.708 | 0.829 | 0.764 | 0.580 | 0.541 | 0.560 | 0.702 | 0.597 | 0.645 | 0.808 | 0.900 | 0.852 |
| Ours-S | 0.719 | 0.744 | 0.842 | 0.790 | 0.612 | 0.583 | 0.597 | 0.704 | 0.600 | 0.648 | 0.814 | 0.913 | 0.861 |
| Ours-A | 0.728 | 0.752 | 0.836 | 0.792 | 0.622 | 0.599 | 0.610 | 0.710 | 0.616 | 0.660 | 0.826 | 0.908 | 0.865 |
| Ours | 0.739 | 0.764 | 0.830 | 0.796 | 0.632 | 0.623 | 0.627 | 0.722 | 0.638 | 0.677 | 0.838 | 0.892 | 0.864 |

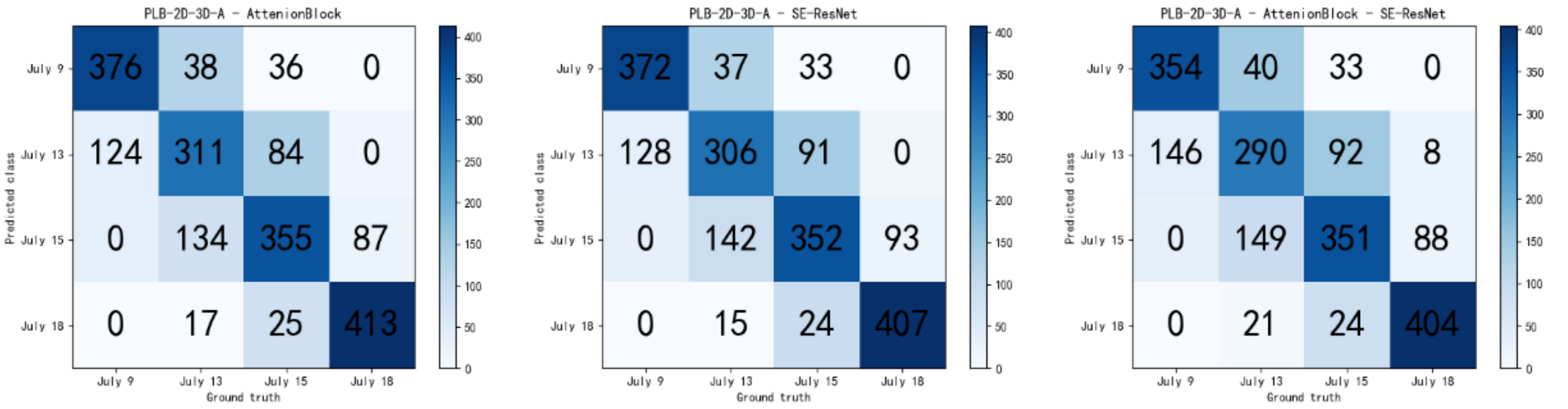

**Fig. 10.** Confusion matrix for ablation experiments at the full bands.

**Table 4.** Ablation experiments at specific bands.

| Models | | July9 | | | July13 | | | July15 | | | July18 | | |
|---|---|---|---|---|---|---|---|---|---|---|---|---|---|
| | Accuracy | precision | recall | F1 | precision | recall | F1 | precision | recall | F1 | precision | recall | F1 |
| Ours-A-S | 0.741 | 0.742 | 0.877 | 0.804 | 0.680 | 0.605 | 0.640 | 0.720 | 0.632 | 0.673 | 0.820 | 0.921 | 0.868 |
| Ours-S | 0.769 | 0.780 | 0.894 | 0.833 | 0.700 | 0.643 | 0.670 | 0.742 | 0.663 | 0.700 | 0.852 | 0.926 | 0.887 |
| Ours-A | 0.779 | 0.786 | 0.889 | 0.834 | 0.704 | 0.663 | 0.683 | 0.760 | 0.677 | 0.716 | 0.866 | 0.929 | 0.895 |
| Ours | 0.790 | 0.796 | 0.888 | 0.839 | 0.720 | 0.679 | 0.699 | 0.764 | 0.700 | 0.731 | 0.878 | 0.922 | 0.899 |

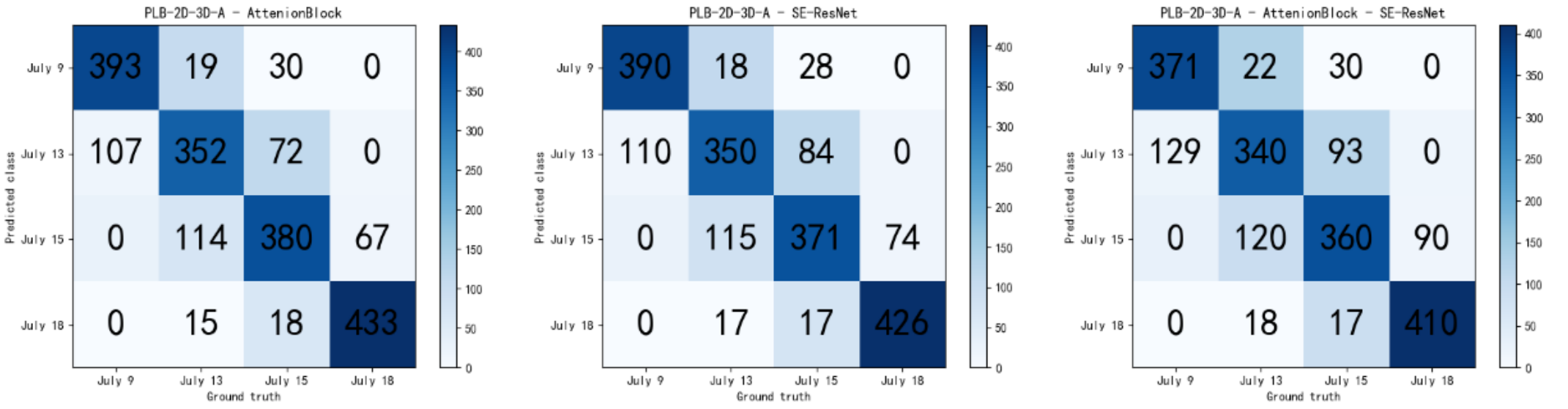

**Fig. 11.** Confusion matrix for ablation experiments at the specific bands.

It can be clearly observed that the attention mechanism has a positive effect on the classification performance. At the full band, removing the AttenionBlock module and the SE-ResNet module reduced the accuracy by 0.011 and 0.02, respectively. At the specific band, removing the AttenionBlock module and the SE-ResNet module reduced the accuracy by 0.011 and 0.021, respectively. It is worth noting that the AttenionBlock module has the most significant ability to distinguish the second class (July13). At specific bands, removing the AttenionBlock module reduced the precision, recall and $F_1$ of the model by 0.016. The SE-ResNet module showed the most striking ability to distinguish the third category (July15). At specific bands, removing the SE-ResNet module reduced the precision, recall and $F_1$ of the model by 0.022, 0.037 and 0.031, respectively.

## 4. Discussion

This paper focuses on early detection of late blight caused by *P. infestans* in potato. It evaluates the combination of 2D-CNN and 3D-CNN with deep synergy. First, PLB-2D-3D-A uses a multilayer fusion strategy to design the network structure. In the horizontal structure, 2D convolution kernel and 3D convolution kernel are used to conduct shallow feature extraction for spectral features and spatial features respectively. In the vertical structure, the general convolution and pooling operations only consider pixels of kernel size and ignore the relationship between the whole feature image pixels, especially the relationship between two pixels that are far away from each other. This could lead to massive entanglement of similar features and seriously affect the classification ability of the model. Therefore, we introduce the AttenionBlock attention mechanism to provide a decoupling-like operation for similar features. However, applying the attention mechanism to the entire feature map results in a severe computational burden. To deal with this, we cropped the size of the original image and used small neighbourhood modules of different sizes. As shown in Table 3 and Table 4, by removing the AttenionBlock attention mechanism from the proposed model structure, the classification accuracy suffers both at full bands and at specific bands. Specially for the second class (13 July), under the specific bands, the precision and recall were reduced by 2.22% and 2.36%, respectively. Due to the large number of sample parameters (over 50 million parameters to be processed for a single sample), we designed the horizontal and vertical structures with fewer convolution operations to ensure efficient network training, resulting in a shallow network depth and width. Generally, it is considered that increasing the depth and width of the network can effectively improve the model accuracy, such as Inception V3 and Resnet101 (Liao et al., 2020). To extract deeper spectral and spatial features, we use the widely accepted residual network as the underlying architecture and add the SE block for HSI classification after the residual transformation. The main role of the SE block is to recalibrate the feature mapping after the residual transformation. The SE block takes the feature map as its input and decomposes the spatial dimensional dependencies by global averaging pooling to learn a channel-oriented descriptor that it passes through a squeeze function. The goal of the descriptor is to emphasise useful channels by recalibrating the feature map, thus embedding the global distribution of feature maps for different channels, and is able to slightly improve the quality of the features generated by the residual block through explicitly modelling the relationship between the channels of its convolutional feature map. As shown in Table 1, when we remove the SE-ResNet module from the proposed model structure, the classification accuracy suffers both at full band and at specific bands. Especially for the third class (July 15), the precision and recall are reduced by 2.88% and 2.36% at the specific band, respectively. It can be seen that the AttenionBlock module and the SE-ResNet module can better fuse spatial and spectral features and have some non-negligible positive effects on the classification of infected samples.

After discussing the PLB-2D-3D-A network structure, we review the whole experimental part. First, to compare the classification effects at the full bands and the specific bands, we used spectral first-order differentiation and spectral second-order differentiation to screen out six specific bands (492nm, 519nm, 560nm, 592nm, 717nm and 765nm) through the intersections of the two differentiation results. We counted the mean values of the spectral curves of infected samples for four days, as shown in Fig. 12. Between 500nm-600nm, the reflectance showed a clear regular variation as the infection level increased, consistent with the results of the band screening. When the band was between 700nm-800nm, the spectral mean reflectance of the second class (July 13) and third class (July 15) showed confusion, especially the spectral curve of the second class (July 13). This is consistent with the results in Tables 1 and 2. At the full bands, the classification accuracy and recall of the second class (July 13) were 0.632 and 0.623, respectively. At the specific bands, the classification accuracy and recall of the second class (July 13) were 0.720 and 0.679, respectively. when we removed the AttenionBlock module and the SE-ResNet module entirely, as shown in Table 3 and Table 4, the classification accuracy and recall of the second class (July 13) decreased by 8.23% and 13.16%, respectively. At the specific bands, the classification accuracy and recall of the second class (July 13) decreased by 5.56% and 10.9%, respectively. At the full band, the classification accuracy and recall of the third class (July 15) decreased by 2.86% and 6.43%, respectively. At the specific band, the classification accuracy and recall of the second class (13 July) sample decreased by 5.76% and 9.71%, respectively. The positive impact of the AttenionBlock module and the SE-ResNet module on HSI classification was verified from a quantitative perspective. An interesting observation is that when removing the AttenionBlock module at full bands, the recall improves by 0.016 and when removing the SE-ResNet module, the recall improves by 0.021. When removing both the AttenionBlock and SE-ResNet modules, the recall improves by 0.008. It should be stated that the goal of this paper is to address the early detection of late blight, and the fourth class of samples already has some spots that could be captured by the experienced human eye, so the fourth class is not considered an early detection of late blight. However, we found that the attention mechanism was not effective in classifying the class with the strongest variation in reflectance, probably because, the features of the fourth class could be extracted well by the traditional convolution mechanism, and embedding the attention mechanism would conversely generate redundant invalid features, thus reducing the generalization ability of the model. When the band is between 800nm-1000nm, the reflectance again shows an extent of regularity. When the band is between 910nm-940nm, the first class (July 9) and the third class (July 15) show a crossover point, indicating that these two classes are prone to feature entanglement. Some studies have shown that when using individual cultivars, there will be some improvement in classification performance in a specific band. When modelling for one cultivar, we need to avoid using feature wavelengths with potentially substantial feature entanglement. It is worth emphasising that by using 20 cultivars and two band screening methods, the final results are intersected and many sensitive bands of cultivars are removed, for example in breeding line 1, 463nm is a specific band and when intersected, this specific band is removed. This will result in a slight decrease in the accuracy of the model. In a realistic scenario, it is not practical to model each cultivar individually and the future trend is to integrate more cultivars for modelling. In general, PLB-2D-3D-A performed the worst classification for the second class and the first class was second only to the fourth class. This indicates that it is possible to detect the early stages of late blight by capturing variations in spectral information on the third day of infection with *P infestans* in a field environment. It is unfair to direct compare the results in this paper with other study since different datasets were used, but it still shows a promising accuracy (0.796) and recall (0.888) values for in-field early recognition (July 9) of PLB with the six selected important bands (492nm, 519nm, 560nm, 592nm, 717nm and 765 nm) based proposed PLB-2D-3D-A model.

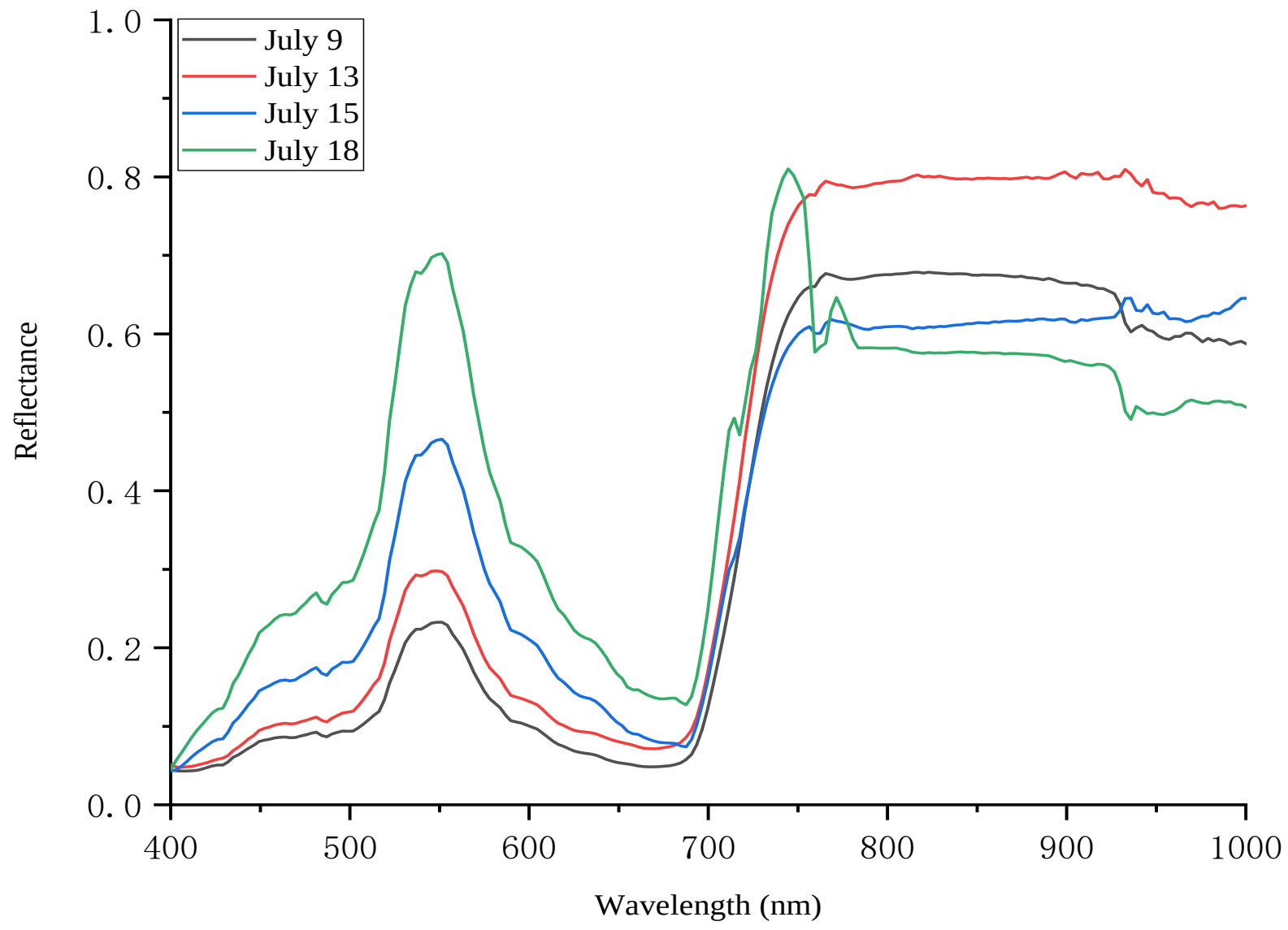

**Fig. 12.** Spectral curves of infected samples for four days.

The red edge is the wavelength between 680 and 750 nm where the slope of the plant reflectance curve is maximum, i.e. the variation in reflectance from the low value region of red light (caused by chlorophyll absorption) to the high value region (leaf and canopy scattering effects). We calculated the red-edge parameter by calculating the first-order differentiation of the spectral reflectance between 680 and 750 nm, and the results are shown in Fig. 13. We found that the peak first-order differential between the 680 and 750 nm bands slowly shifted towards the longer wavelengths with the severity of the infection level, and a red-edge displacement phenomenon occurred. Existing studies have found that changes in chlorophyll and nitrogen with the health of the vegetation can cause red-edge displacement (Molina-Bolivar et al., 2019) phenomena in the spectrum. The occurrence of red-edge shifts in early infected potato leaves also validates from a spectrochemical point of view that it is possible to achieve early late blight detection.

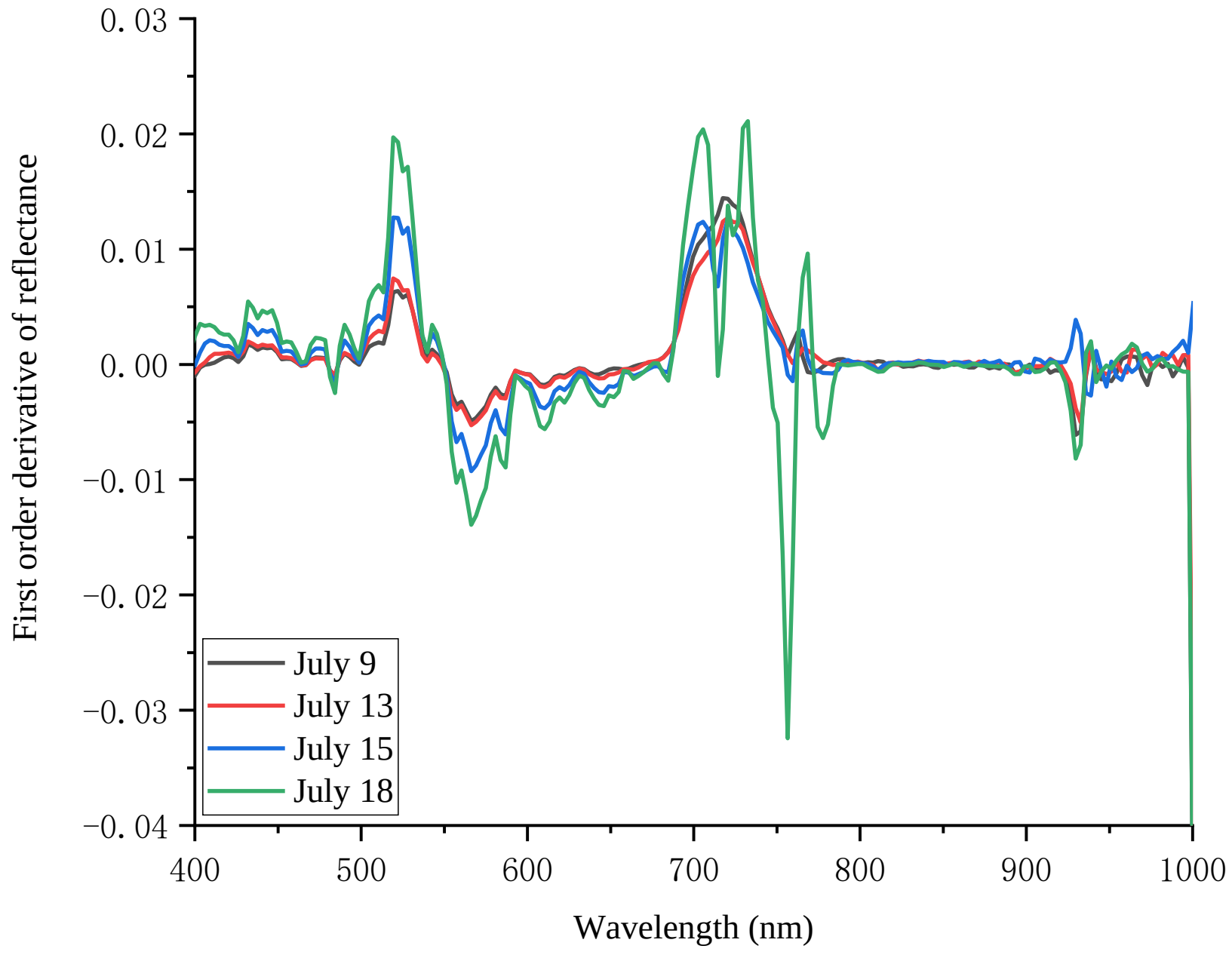

**Fig. 13.** Mapping relationship between wavelength and reflectance first order differentiation.

The PLB-2D-3D-A model shows some encouraging results, but is not convincing enough. In future work, further improvements in model accuracy, particularly for class 2 and class 3, is an urgent matter that needs to be addressed. Not only that, the dataset in this paper does not involve healthy samples, and distinguishing healthy samples from early infected samples is a worthwhile task to explore. Furthermore, Large area canopy level detection can significantly improve operational efficiency in real-world scenarios, whereas this paper focuses on caopy level detection at the single plant level. Therefore, it is also meaningful to establish an accurate geometric-optical model between canopy and leaf spectra. Finally, the emergence of the red-edge displacement phenomenon has brought us inspiration. Currently our work is mainly focused on the early detection of late blight. It is worth exploring the possibility of adding factors to the deep learning inputs that could significantly influence the variations in spectral reflectance, such as variations in chlorophyll and nitrogen content, to further improve the model by incorporating quantitative conditioning mechanisms in the model.

## 5. Conclusions

This paper proposes a deep learning architecture combining 2D-CNN and 3D-CNN, called PLB-2D-3D-A, for accurate early detection of PLB. We collected the original dataset of 20 potato genotypes with 15360 (64x64x204) images. The proposed model achieved an accuracy of 0.739 at the full bands in the test dataset. The accuracy and recall on day 12 after inoculation were the highest, reaching 0.838 and 0.892, respectively. The accuracy and recall on day 3 were second only to the day 12, reaching 0.764 and 0.830, respectively. The accuracy and recall on day 7 were the lowest, reaching 0.632 and 0.623, respectively. The proposed model achieves an accuracy of 0.790 at the specific bands (492, 519, 560, 592, 717 and 765). The accuracy and recall on day 12 are the highest, reaching 0.878 and 0.922, respectively. The accuracy and recall on day 3 are second only to the day 12, reaching 0.796 and 0.888, respectively. The accuracy and recall on day 7 are the lowest, reaching 0.720 and 0.679, respectively. The results show that both in full bands and in specific bands, the proposed model outperforms the accuracy of the traditional machine learning method RF and the deep learning-based methods 2D-CNN and 3D-CNN. The study shows our deep learning-based model achieves a promising result for early in-field detection of PLB based on proximal hyperspectral images.

## CRediT authorship contribution statement

**Chao Qi:** Conceptualization, Methodology, Software, Writing - original draft, Writing - review & editing. **Murilo Sandroni:** Data collection - review & editing. **Jesper Cairo Westergaard:** Data collection - review & editing. **Ea Høegh Riis Sundmark:** Resources, review &editing . **Merethe Bagge:** Resources, review &editing. **Erik Alexandersson:** Supervision, Project administration, Funding acquisition, Writing - review & editing. **Junfeng Gao:** Supervision, Methodology, Writing - review & editing Declaration of Competing Interest.

The authors declare that they do not have any financial or non-financial conflict of interests.

## Acknowledgments

This research was founded by the Nordic Council of Ministers, Copenhagen, Denmark (PPP #6P2 and #6P3) and NordForsk, Norway (#84597). This work was partially supported by Lincoln Agri-Robotics as part of the Expanding Excellence in England (E3) Programme.